\documentclass{article}

\usepackage[utf8]{inputenc}
\usepackage[T1]{fontenc}
\usepackage{cite}
\usepackage[ruled]{algorithm2e}
\usepackage{graphicx}
\usepackage{float}

\usepackage{PRIMEarxiv}

\usepackage[utf8]{inputenc} 
\usepackage[T1]{fontenc}    
\usepackage{hyperref}       
\usepackage{url}            
\usepackage{booktabs}       
\usepackage{amsfonts}       
\usepackage{nicefrac}       
\usepackage{microtype}      
\usepackage{lipsum}
\usepackage{fancyhdr}       
\usepackage{graphicx}       

\usepackage{amsmath}
\usepackage{amssymb}

\usepackage{algorithm2e}

\usepackage{tikz} \usetikzlibrary{positioning, shapes.geometric, arrows.meta}

\usepackage{authblk}

\graphicspath{{media/}}     

\newcounter{pr}

\title{Conditional Denoising Model as a Physical Surrogate Model  
}

\author[1]{\textbf{José Afonso}\thanks{Corresponding author: josefafonso@tecnico.ulisboa.pt}\hspace{0.9ex}}
\author[1]{\textbf{Pedro Viegas}}
\author[2]{\textbf{Rodrigo Ventura}}
\author[1]{\textbf{Vasco Guerra}}
\affil[1]{Instituto de Plasmas e Fusão Nuclear, Instituto Superior Técnico, Av. Rovisco Pais 1, Lisbon, Portugal}
\affil[2]{Instituto de Sistemas e Robotica, Instituto Superior Técnico, Av. Rovisco Pais 1, Lisbon, Portugal}

\begin{document}
\maketitle

\begin{abstract}
Surrogate modeling for complex physical systems typically faces a trade-off between data-fitting accuracy and physical consistency.
Physics-consistent approaches typically treat physical laws as soft constraints within the loss function, a strategy that frequently fails to guarantee strict adherence to the governing equations, or rely on post-processing corrections that do not intrinsically learn the underlying solution geometry.
To address these limitations, we introduce the {Conditional Denoising Model (CDM)}, a generative model designed to learn the geometry of the physical manifold itself.
By training the network to restore clean states from noisy ones, the model learns a vector field that points continuously towards the valid solution subspace.
We introduce a time-independent formulation that transforms inference into a deterministic fixed-point iteration, effectively projecting noisy approximations onto the equilibrium manifold.
Validated on a low-temperature plasma physics and chemistry benchmark, the CDM achieves higher parameter and data efficiency than physics-consistent baselines.
Crucially, we demonstrate that the denoising objective acts as a powerful implicit regularizer: despite never seeing the governing equations during training, the model adheres to physical constraints more strictly than baselines trained with explicit physics losses.

\end{abstract}


\section{Introduction}
\label{sec:Introduction}

Generative models are a class of deep learning algorithms designed to estimate and sample from an unknown data distribution. The remarkable advances in this field, particularly for image generation, have been largely driven by the scalable and stable training of diffusion-based and auto-encoding models \cite{Sohl-Dickstein2015, Ho2020, Song2020}. 
These models excel at capturing the global, intrinsic structure of a data manifold. However, their success is fundamentally reliant on massive datasets, making them challenging to apply in specialized scientific domains where data is often scarce and expensive to generate.

In various domains of physics and engineering, computational simulations are indispensable for understanding complex phenomena. These processes are typically modeled by solving systems of differential equations derived from physical laws. Frequently, the primary interest lies in the steady-state solution, which formally requires finding the fixed point $y$ of a non-linear system of algebraic equations: \begin{equation}
    F(x, y; \theta) = 0, 
    \label{eq:F_equation}
\end{equation} 
where $F$ represents the system of governing equations. In this general formulation, $x$ denotes the experimental control inputs, $y$ represents the physical state of the system, and $\theta$ encapsulates the physical hyperparameters. Solving this system repeatedly for vast parameter spaces is often computationally intractable, necessitating the development of fast and accurate surrogate models.

 In this work, we focus on Low-Temperature Plasma (LTP) physical system as a representative case study. Specifically, we utilize the LisbOn KInetics Boltzmann+Chemistry (LoKI-B+C) simulation tool \cite{Tejero-del-Caz2019, Tejero-del-Caz_2021, Dias_2023} hereafter referred to as LoKI, a kinetic simulator that models the plasma mixture (e.g., \cite{Dias_2023, Viegas2024}). In this specific context, the input $x$ corresponds to operating conditions, such as gas pressure and reduced electric field. The state $y$ comprises the densities of various particle species (charged and neutral), and $\theta$ corresponds to physical hyperparameters, including rate constants and transport coefficients.

The surrogate modeling task faces a fundamental obstacle. In many complex simulation domains, including simulators like LoKI, the full algebraic system $F$ is not readily accessible as a differentiable function for training. The simulator often acts as a \textit{black-box} solver, providing only the final steady-state solutions $y$ for a given set of inputs $x$. This practical constraint makes the direct implementation of methods that require explicit access to $F$, such as traditional, residual-based Physics-Informed Neural Networks (PINNs) \cite{Raissi2019} or methods that project onto the constraint manifold \cite{Jacobsen2025}, computationally infeasible.
Furthermore, even if $F$ were accessible, residual-based physics-informed paradigms suffer from pathologies related to conflicting gradients between data-fitting and physics-penalty terms \cite{Wang2022}.

Given the inaccessibility of $F$, standard neural networks act as physics-agnostic regressors $(f(x)\approx y)$, often yielding physically meaningless results. Recent alternatives attempt to substitute $F$ with general physical invariants (e.g., conservation laws) \cite{Valente2025}. However, Valente et al. \cite{Valente2025} showed that incorporating these constraints into the loss function is often ineffective. Their proposed output projection successfully enforces validity but operates strictly as a post-hoc correction and hence does not learn the true physical data manifold.

To bypass these constraints, we seek a data-driven paradigm capable of implicitly learning the geometry of the physical manifold $F(x, y; \theta) = 0$. Score-based and diffusion models \cite{Sohl-Dickstein2015, Ho2020, Song2020} offer a particularly promising direction, as they are designed to model complex distributions by capturing the underlying manifold structure. In physics, these models have been applied as generative solvers for partial differential equations (PDEs) \cite{Jacobsen2025, Hong2024, Park2025} or for modeling molecular dynamics \cite{Hsu2024, Wang2025}. 
Regarding their conditional variants \cite{Ho2022}, they are traditionally applied to inverse problems \cite{Hong2024, Song2022} or time-dependent dynamics \cite{Ho2022video, Yuan2023}, and they are typically demonstrated on large-scale datasets. However, recent works \cite{Kang2024, cheng2025physicsinformedconditionaldiffusionmodel} have begun to demonstrate their efficacy as direct physical surrogates, validating their potential for
sample-efficient learning in data-scarce scientific domains.

In this work, we construct conditional surrogate models for deterministic steady-state simulators, effectively modeling a Dirac delta conditional distribution $p(y|x)$. We propose the Conditional Denoising Model (CDM), which generalizes Conditional Denoising Autoencoders \cite{Viencent2008} to a continuous noise spectrum, bridging the geometric intuition of denoising autoencoders with the objective of  diffusion models \cite{Ho2020}.
 We introduce a \textit{time-independent} formulation of the CDM that learns a static vector field continuously pointing toward the solution manifold. By removing the time-dependency typical of diffusion models, we transform inference into a deterministic, iterative fixed-point refinement. 
We evaluate the CDM on the LoKI kinetic simulator for Low-Temperature Plasma. Benchmarking against physics-consistent  regressors demonstrates that our method robustly captures complex physical correlations, even in low-data regimes.

The paper is organized as follows: Section \ref{sec:Background} reviews the background on manifold learning and diffusion models. Section \ref{sec:Methodology} details the CDM framework and the proposed deterministic inference scheme (CDM-0). Section \ref{sec:Experiments} evaluates the method on the LoKI plasma dataset, and Section \ref{sec:Conclusions} concludes.

\section{Background}
\label{sec:Background}

\subsection{Generative Modeling via Denoising}

Denoising Diffusion Probabilistic Models (DDPMs) \cite{Ho2020} generate data by reversing a gradual, intentional corruption process. Given a distribution of original, uncorrupted data $x_0 \sim p_{\text{d}}(x)$ (referred to as \textit{clean}), a forward diffusion process systematically adds Gaussian noise over time $t \in [0, T]$, producing a sequence of latent variables $x_t$. The standard training objective involves learning a network $\epsilon_\theta(x_t, t)$ to predict the noise component $\epsilon \sim \mathcal{N}(0, I)$ (where $I$ denotes the identity matrix) that was added to the signal. This is typically formulated as minimizing the variational lower bound (ELBO) on the negative log-likelihood \cite{Ho2020, kingma2023variationaldiffusionmodels}:
\begin{equation}
    \mathcal{L}_{\text{Diff}} = \mathbb{E}_{t, x_0, \epsilon} \left[ w(t) \| \epsilon_\theta(x_t, t) - \epsilon \|^2 \right],
\end{equation}
where $\omega(t)$ corresponds to the weighting function.

In the unconditional setting, inference is performed by iteratively subtracting the predicted noise to traverse the reverse Markov chain from $p(x_T)$ back to $p(x_0)$ \cite{Ho2020, Song2020}.

While this $\epsilon$-prediction formulation has popularized diffusion models for large-scale image synthesis, recent theoretical works suggest it may be suboptimal for data governed by strict low-dimensional constraints \cite{li2025basicsletdenoisinggenerative}. We explicitly address this limitation in our proposed method, as detailed below.

\subsection{The Physical Manifold Assumption}

A fundamental consequence of the governing equation $F(x, y; \theta) = 0$ (Eq.\eqref{eq:F_equation}) is that valid physical solutions $y$ do not occupy the entire $D$-dimensional ambient output space $\mathbb{R}^D$.
Instead, they are constrained to a lower-dimensional, highly structured subspace $\mathcal{M} \subset \mathbb{R}^D$, commonly referred to as the data manifold~\cite{Cayton2005, Narayanan2010}.
In the context of physical surrogates, this structure is rigorous rather than hypothetical: the manifold is explicitly defined as the set of points satisfying the governing equations.
This geometric constraint creates a sharp distinction between the physical signal and auxiliary noise.
As analyzed by Li et al.~\cite{li2025basicsletdenoisinggenerative}, the clean physical state $y$ lies strictly on $\mathcal{M}$, whereas noise $\epsilon$ is inherently distributed across the full high-dimensional observation space~\cite{Bengio2013}.
Consequently, training a network to predict noisy or unstructured quantities wastes model capacity on the high-entropy ambient space.

In contrast, targeting the clean data allows the model to act as a projection operator, focusing solely on the low-dimensional manifold geometry.
This distinction is important for our approach: given the scarcity of scientific data, focusing the learning process on the manifold structure maximizes data efficiency and ensures the surrogate adheres to the underlying physical system rather than fitting orthogonal noise components \cite{li2025basicsletdenoisinggenerative}.

\subsection{Manifold Learning via Denoising Autoencoders}
\label{sec:Manifold_DAE}
To explicitly 
constrain the surrogate model to the underlying data manifold, we adopt the Conditional Denoising Autoencoder (CDAE) framework \cite{Viencent2008}. Unlike standard diffusion implementations that target the noise $\epsilon$, the CDAE is designed to learn the geometry of the physical manifold itself by learning a restoration vector field \cite{Alain2014}.



The training procedure mirrors the manifold assumption: we start with a valid physical state $y \in \mathcal{M}$ and corrupt it with isotropic Gaussian noise of intensity $\sigma$ to produce $\tilde{y} \sim q_\sigma(\cdot | y) = \mathcal{N}(\tilde{y}; y, \sigma^2 I)$.
Geometrically, this perturbation pushes the data point off the manifold into the surrounding ambient space~\cite{Bengio2013, Alain2014}.
The network $g_\phi$ is then tasked with reversing this perturbation.
By minimizing the reconstruction error between the prediction and the clean signal:
\begin{equation}
    \mathcal{L}_{\text{DAE}}(\phi) = \mathbb{E}_{(x, y) \sim p_d} \mathbb{E}_{\tilde{y} \sim q_\sigma(\cdot|y)} \left[ \| g_\phi(\tilde{y}, x) - y \|^2_2 \right],
    \label{eq:DAE_objective}
\end{equation}
where $p_d$ is the data distribution. This way, the model learns a vector field that effectively projects any point in the vicinity of the manifold back onto it. 

Crucially, it has been shown that minimizing the denoising objective Eq.\eqref{eq:DAE_objective} is mathematically equivalent to \textit{Score Matching} \cite{Vincent2011, Hyvarinen2005}. The optimal denoiser $g_{\phi^*}$ implicitly learns the score function (the gradient of the log-density) of the data distribution \cite{Song2020, Vincent2011}. This theoretical connection bridges the gap between denoising autoencoders and score-based generative models, allowing us to leverage the CDAE as a suitable framework for learning the \textit{fixed-point} solutions of physical systems.

\section{Methodology: Conditional Denoising Model (CDM)}
\label{sec:Methodology}

\subsection{Training Objective: Multi-Scale Manifold Learning}

While the classical Denoising Autoencoder (DAE) introduced in Section~\ref{sec:Manifold_DAE} provides a mechanism for manifold projection, training at a single noise intensity $\sigma$ is insufficient for a full generative model.
If $\sigma$ is too small, the model only learns local manifold corrections, failing to capture the global topology of the solution space~\cite{Song2020}.
Conversely, if $\sigma$ is too large, the model learns the coarse global structure but loses fine-grained physical details~\cite{Viencent2008, Bengio2013}.

To resolve this trade-off between global coherence and local precision, we formulate our Conditional Denoising Model (CDM) as the conditional counterpart to the \textit{$x$-prediction} parameterization found in diffusion and score-based models~\cite{Ho2020, Song2020, li2025basicsletdenoisinggenerative, Salimans2022}.
It extends the autoencoding objective across a continuous spectrum of noise levels $\sigma(t)$.
Following the Variance Exploding (VE) formulation~\cite{Song2020}, the time variable $t$ defines the noise intensity $\sigma(t)$, monotonically mapping the interval from a minimum noise level $\sigma_{\min} \approx 0$ (at $t=0$) to a maximum $\sigma_{\max}$ (at $t=1$).
This parameterization allows the network to learn the manifold structure hierarchically, resolving first the global geometry at high noise levels and refining local physical constraints as the noise vanishes~\cite{Ho2020}.

The training objective minimizes the weighted reconstruction error across the entire noise schedule:
\begin{equation}
    \mathcal{L}_{\rm CDM}(\phi) = \mathbb{E}_{t\sim \mathcal{U}(0,1)} \, \mathbb{E}_{(x,y) \sim p_{\rm d}}  \mathbb{E}_{\tilde{y} \sim q_{t} (\cdot | y)} \left[ \omega(t) \| g_{(t),\phi}(\tilde{y}, x) - y\|^2_2\right],
    \label{eq:CDM_train_objective}
\end{equation}
where $\mathcal{U}$ represents the uniform distribution and $g_{(t),\phi}$ corresponds to the neural network with explicit or implicit time dependency. It is trained to recover the clean physical state $y$ given the noisy input $\tilde{y}$ and experimental conditions $x$, while using $t$ to index the current noise intensity $\sigma(t)$.
Consistent with the VE schedule, the forward corruption kernel is given by $q_{t}(\tilde{y} | y) = \mathcal{N}(\tilde{y}; y, \sigma(t)^2 I)$.
The complete training procedure is summarized in algorithm~\ref{alg:PCDAE_Training}.

\subsection{Theoretical Foundations}
We adopt the standard weighting $\omega(t)=1$ \cite{Ho2020}. Under this configuration, our simple clean-data prediction objective is supported by three complementary theoretical frameworks that justify its use as a generative model.

\textbf{Variational and Recovery Perspectives.} 
First, the objective can be viewed through the lens of variational inference. Since the clean data $y$ and noise $\epsilon$ are linearly related, predicting $y$ is mathematically equivalent to predicting $\epsilon$ up to a scaling factor. Thus, optimizing Eq.\eqref{eq:CDM_train_objective} effectively maximizes the  ELBO objective, forcing the model to approximate the true data distribution by minimizing the accumulated denoising error \cite{Ho2020}. 
Simultaneously, this objective maximizes the expected \textit{recovery log-likelihood} $\log p_\phi(y|\tilde{y}, x)$ \cite{Bengio2013, Gao2021}. By parameterizing the recovery distribution as an isotropic Gaussian centered at the model prediction, $p_\phi(y | \tilde{y}, x) = \mathcal{N}(y; g_{\phi}(\tilde{y}, x, t), \sigma^2(t) I)$, maximizing the likelihood becomes identical to minimizing the MSE:
\begin{equation}
    \max_{\phi} \mathbb{E}[\log p_\phi(y | \tilde{y}, x)] \iff \min_{\phi} \mathbb{E} \left[ \frac{1}{2 \sigma^2(t)}\| g_{\phi}(\tilde{y}, x, t)- y \|^2_2 \right].
\end{equation}
This ensures the model effectively \textit{recovers} the clean manifold geometry from any point in the ambient space.

\textbf{Distributional Matching via DCD.}
To quantify the alignment between the modeled and true distributions, we analyze the objective using the \textit{Diffusion Contrastive Divergence} (DCD) framework \cite{luo2023}. Adapting the formulation to our conditional setting (see derivation in Appendix \ref{app:DCD_derivation}), our loss function minimizes the following conditional DCD:
\begin{equation}
    \min_\phi \mathbb{E}_{t\sim \mathcal{U}(0,1)} \, \mathbb{E}_{x\sim p_d(x)} \left[ \mathcal{D}^{(t)}_{\rm DCD} (p_d(y|x) ||p_\phi(y|x)) \right],
    \label{eq:DCD_conditional_t}
\end{equation}
where $\mathcal{D}^{(t)}_{\rm DCD}$ is a valid divergence defined as the difference between the KL divergences of the clean and perturbed distributions:
\begin{equation}
    \mathcal{D}^{(t)}_{\rm DCD} (p_d || p_\phi) = \mathbb{E}_{p_d(x)} \left[ \mathcal{D}_{\rm KL}(p_d || p_\phi) - \mathcal{D}_{\rm KL}(p^{(t)}_d || p^{(t)}_\phi)\right].
\end{equation}
The terms $p^{(t)}_d$ and $p^{(t)}_\phi$ represent the true and model distributions, respectively, after being perturbed by the forward diffusion process $q_t$:
\begin{equation}
    p^{(t)}(\tilde{y} | x) = \int dy \,  q_t(\tilde{y} | y) p(y, x),
\end{equation}
where the forward diffusion kernel $q_t(\tilde{y}|y)$ follows the VE schedule.

Since $\mathcal{D}^{(t)}_{\rm DCD} \geq 0$ and equals zero if and only if $p_d(y|x) = p_\phi(y|x)$ \cite{luo2023}, minimizing Eq.\eqref{eq:DCD_conditional_t} guarantees that our surrogate model converges to the true conditional distribution of the physical system i.e, to the true physical solution.

\textbf{Score Matching.}
Finally, the learned map $g_{\phi}$ has a direct connection to the score function of the data distribution. By the conditional \textit{Tweedie}'s formula \cite{Park2025_tweedie}, the optimal MSE denoiser $g_{\phi^*}$ is related to the score of the noisy density $\nabla_{\tilde{y}} \log p^{(t)}_d(\tilde{y} | x)$ via:
\begin{equation}
    g_{\phi^*}(\tilde{y}, x, t) = \mathbb{E}[y | \tilde{y}, x, t] =  \tilde{y} +  \sigma^2(t)  \nabla_{\tilde{y}} \log p^{(t)}_d(\tilde{y} | x).
    \label{eq:Conditional_Tweedies_formula}
\end{equation}
For completeness, the full derivation is presented in Appendix \ref{app:vector_field_and_conditional_score}.
This confirms that minimizing Eq.\eqref{eq:CDM_train_objective} is an implementation of Denoising Score Matching \cite{Vincent2011}. Accordingly, it ensures that our model learns the gradient vector field required to guide the reverse diffusion process from noise back to the physical manifold.

\subsection{Inference as a Generative Flow}

Inference is the process of generating a physical state $y$ for a given condition $x$ by reversing the diffusion chain.
Instead of relying on random steps to find the solution, our goal is to use a deterministic path that leads from the initial noise back to the valid physical manifold.
To achieve this, we formulate the inference as a \textit{Generative Flow} governed by a Probability Flow ODE~\cite{Song2021}, effectively replacing the random noise injection of stochastic methods with a smooth, deterministic trajectory.

\subsubsection{Derivation of the Inference ODE}
The Variance Exploding (VE) corruption process utilized in our training objective (Eq.~\ref{eq:CDM_train_objective}) corresponds to the following Stochastic Differential Equation (SDE)~\cite{Song2021}:
\begin{equation}
    d\tilde{y} = \sqrt{\frac{d[\sigma^2(t)]}{dt}} d\omega,
\end{equation}
where $d\omega$ denotes a standard Wiener process \cite{Song2021}.
To generate samples, we reverse this dynamics using the Probability Flow ODE:
\begin{equation}
    d\tilde{y} = -\frac{1}{2} \frac{d[\sigma^2(t)]}{dt} \nabla_{\tilde{y}} \log p_t(\tilde{y}|x) dt.
\end{equation}
Leveraging the score approximation inherent in our model via the conditional Tweedie's formula, we substitute the score term $\nabla_{\tilde{y}} \log p_t \approx (g_\phi - \tilde{y})/\sigma^2(t)$ to derive the specific inference dynamics for our Conditional Denoising Model:
\begin{equation}
    d\tilde{y} = \frac{1}{\sigma(t)} \frac{d\sigma(t)}{dt} (\tilde{y} - g_\phi(\tilde{y}, x, t)) dt.
    \label{eq:final_ode}
\end{equation}
This equation reveals that the generative flow is driven strictly by the \textit{restoration vector} $(g_\phi - \tilde{y})$, scaled by the logarithmic rate of change of the noise schedule.

\subsubsection{Discrete Sampling Algorithm}
To solve Eq. \eqref{eq:final_ode} numerically, we discretize the time steps $i=N, N-1, \dots, 0$, corresponding to a noise schedule $\sigma_{N} > \dots > \sigma_{0} \approx 0$. Integrating the ODE over a step from $\sigma_i$ to $\sigma_{i-1}$ using a first-order Euler approximation \cite{Karras2022}, we obtain the iterative update rule:
\begin{equation}
    y_{i-1} = y_i + \frac{\sigma_i - \sigma_{i-1}}{\sigma_i} (g_\phi(y_i, x, \sigma_i) - y_i).
\end{equation}
Defining the relative step size $\eta_i = (\sigma_i - \sigma_{i-1})/\sigma_i$, where $\eta_i \in [0, 1]$, this formulation simplifies the diffusion sampling process into an intuitive iterative refinement:
\begin{equation}
    y \leftarrow y + \eta_i \, (g_\phi(y, x, \sigma) - y).
    \label{eq:update_rule}
\end{equation}

This update rule interprets generation as a sequence of \textit{partial projections}: at each step, the model predicts the projection onto the clean manifold $g_\phi$, and the sampler moves the current state $y$ a fraction $\eta_i$ along that vector. 

\subsubsection{Architectural Variants: Time-Dependency}
The nature of the learned vector field depends on whether the network $g_{(t),\phi}$ is explicitly conditioned on the noise level $t$. We explore two distinct design choices:

\paragraph{Explicit Time-Dependent Vector Field (CDM-$t$)}
In the standard diffusion formulation, the network $g_{\phi}(y, x, t)$ takes $t$ (or equivalently $\sigma$) as an explicit input. The network learns a \textit{dynamic} vector field that evolves as the inference trajectory proceeds. This corresponds to the classical ODE solver approach \cite{Song2020, Karras2022}. 
Algorithm \ref{alg:PCDAE_t_Inference} summarizes this procedure.

\paragraph{Implicit Time-Independent Vector Field (CDM-$0$)}
Alternatively, we consider a time-independent parameterization $g_\phi(y, x)$, where the network is blind to the noise level $\sigma(t)$, as implemented in algorithm~\ref{alg:PCDAE_0_Inference}.
In this regime, the model learns a \textit{static} vector field pointing continuously towards the manifold.
Since the network is trained via $\mathbb{E}_{t}$ (Eq.~\eqref{eq:CDM_train_objective}) to denoise across all perturbation scales, it learns a global contraction map valid over the entire domain $[\sigma(0), \sigma(1)]$.

This independence allows us to define the effective \textit{time} axis during inference purely through the step size $\eta$.
From our derived update rule (Eq.~\eqref{eq:update_rule}), choosing a constant step size $\eta_i = \eta$ implicitly defines a geometric decay schedule, where $\sigma_k = \sigma_{\text{start}} (1 - \eta)^k$.
Under this scheme, the inference process simplifies to a \textit{Fixed Point Iteration}:
\begin{equation}
    y_{k+1} = y_k + \eta \, (g_\phi(y_k, x) - y_k),
\end{equation}
which is analogous to performing gradient descent on an implicit energy landscape~\cite{Wang2025_EqM, Grathwohl2020}.
Unlike the time-dependent case, CDM-$0$ relies on convergence criteria: the iteration terminates when the residual norm $\|g_\phi(y_k, x) - y_k\|_2$ falls below a tolerance $\epsilon_{\text{conv}}$, indicating that the state has settled on the physical manifold.

In this work, we evaluate two step-size strategies for this fixed-point iteration.
First, we use the constant step size described above.
Second, to better handle the varying scales of physical quantities, we introduce an adaptive step size.
While a constant $\eta$ implies a standard geometric decay of the noise, this adaptive rate scales with the relative error:
\begin{equation}
    \eta_k = \eta_{\text{base}} \cdot \frac{\|g_\phi(y_k, x) - y_k\|_1}{\|y_k\|_1 + \delta}.
\end{equation}
This formulation allows the solver to take larger steps when far from the solution and finer, more conservative steps as it approaches equilibrium, ensuring convergence across different physical regimes without manual schedule tuning.

\begin{figure}[H]
    \begin{minipage}[t]{0.48\textwidth}
        \begin{algorithm}[H]
            \SetAlgoLined
            \DontPrintSemicolon
            \SetKwInOut{Input}{Input}
            \SetKwInOut{Output}{Output}
            
            \Input{Data $p_d(x,y)$, schedule $\sigma(t)$, network ${\phi}$ (time-dep. or time-indep.), \textit{time-dependent} flag}
            \Output{Trained network $\phi$}

            \While{not converged}{
                $\triangleright$ \textit{Data Sampling}:\\
                $x, y \sim p_{d}(x, y), \quad$  $z \gets [y, x]^T, \quad$ $t \sim \mathcal{U}(0,1),$ \; 
                
                $\triangleright$ \textit{Noise Injection}:\\
                $\sigma \gets \sigma(t)$ \;
                $\tilde{y} \sim \mathcal{N}(y, \sigma^2 I)$ \;
                
                $\triangleright$ \textit{Compute Loss}:\\
                \eIf{$g_\phi$ is time-dependent}{
                    $z_{pred} \gets g_{\phi}(\tilde{y}, x; \sigma)$
                }
                {
                    $z_{pred} \gets g_{\phi}(\tilde{y}, x)$
                }
                
                $\mathcal{L} \gets \frac{1}{B}\sum^B_{i=1} \| z_{pred} - z_i \|^2_2$ \;
                
                $\triangleright$ \textit{Update}:\\
                $\Delta \phi \leftarrow \nabla_\phi \mathcal{L}$ \;
                Apply Adam update using $\Delta \phi$
            }
            \caption{CDM Training}
            \label{alg:PCDAE_Training}
        \end{algorithm}
    \end{minipage}
    \hfill
    \begin{minipage}[t]{0.48\textwidth}
\begin{algorithm}[H]
    \footnotesize
    \SetAlgoLined
    \DontPrintSemicolon
    \SetKwInOut{Input}{Input}
    \Input{Condition $x$, Model $g_{\phi}$, Noise Schedule $\{\sigma_i\}_{i=N}^0$ (where $\sigma_N > \dots > \sigma_0 \approx 0$), Steps $K$ (optional refinement), Clip $\epsilon_{\text{max}}$}
    
    Initialize $y_N \sim \mathcal{N}(0, I)$ \;
    
    \For{$i = N$ \textbf{to} $1$}{
        $\triangleright$ \textit{Calculate dynamic Euler step size based on schedule}:\\
        $\sigma_{\text{curr}} \leftarrow \sigma_i, \quad \sigma_{\text{next}} \leftarrow \sigma_{i-1},$ \;
        $\eta_i \leftarrow (\sigma_{\text{curr}} - \sigma_{\text{next}}) / \sigma_{\text{curr}},$ \;
        
        \For{$k=1$ \textbf{to} $K$}{
            $\triangleright$ \textit{Predict clean data conditioned on time $t_i$}:\\
            $\hat{y} \leftarrow g_{\phi}(y_i, x, \sigma_{\text{curr}})$ \;

            $\triangleright$ \textit{Compute displacement vector (residual)}:\\
            $v \leftarrow \hat{y} - y_i$ \;

            $\triangleright$ \textit{Euler Update: Move towards clean manifold}:\\
            $y_i \leftarrow y_i + \eta_i \, v$ \;
        }
        $y_{i-1} \leftarrow y_i$ \;
    }
    \Return $y_0$
    \caption{CDM-$t$ Inference}
    \label{alg:PCDAE_t_Inference}
\end{algorithm}
        
        \vspace{0.2cm} 
        
\begin{algorithm}[H]
    \footnotesize 
    \SetAlgoLined
    \DontPrintSemicolon
    \SetKwInOut{Input}{Input}
    \Input{Condition $x$, Model $g_{\phi}$, Base Rate $\eta$, Max Steps $N_{\max}$, Tol $\epsilon_{\text{conv}}$, Flag \textit{Adaptive}}
    
    Initialize $y_0 \sim \mathcal{N}(0, I)$ \;
    
    \For{$k=0$ \textbf{to} $N_{\max}$}{
        $\triangleright$ \textit{Predict clean manifold projection}:\\
        $\hat{y} \leftarrow g_{\phi}(y_k, x)$ \;
        
        $\triangleright$ \textit{Compute displacement vector (Residual)}:\\
        $v \leftarrow \hat{y} - y_k$ \;

        $\triangleright$ \textit{Check for convergence (Equilibrium reached)}:\\
        \If{$\| v \|_2 < \epsilon_{\text{conv}}$}{
            \textbf{break} \;
        }

        $\triangleright$ \textit{Optional: Adaptive Step Sizing}:\\
        \eIf{\text{Adaptive}}{
            $\eta_k \leftarrow \eta \cdot \frac{\|v\|_1}{\|y_k\|_1 + \delta} \quad $            $\triangleright$ \textit{Scale by relative error}
        }{
            $\eta_k \leftarrow \eta \quad$
            $\triangleright$ \textit{Standard geometric decay}
        }

        $\triangleright$ \textit{Fixed Point Update}: \\
        $y_{k+1} \leftarrow y_k + \eta_k \, v$ \;
    }
    \Return $y_{k+1}$
    \caption{CDM-$0$ Inference }
    \label{alg:PCDAE_0_Inference}
\end{algorithm}
    \end{minipage}
\end{figure}

\subsection{Justification of the Denoising Objective}

The choice of the denoising objective (Eq.\eqref{eq:CDM_train_objective}) is motivated by both its information-theoretic properties and its superior statistical efficiency in our problem domain. For the following discussion, let $Y_1$ and $Y_0$ be random variables representing a sample from the true data distribution $p_d(y) = \int  p_d(x, y) dx$ and the isotropic noise distribution $\mathcal{N}(0, I)$, respectively. 

From an Information Bottleneck (IB) perspective \cite{Tishby2000}, our objective is a principled generalization of standard regression. A typical regressor learns $\mathbb{E}[Y_1|X]$. Our model learns the more general quantity $\mathbb{E}[Y_1| X,Y_t]$, where $Y_t = Y_1 + \sigma(t) Y_0$ contains information about $Y_1$ that is continuously controlled by the noise level $\sigma(t)$ with $t \in [0, 1]$. Training across this spectrum forces the model to learn a robust, multi-scale representation. In the high-noise limit $(t \to 1)$, the mutual information $I(Y_1;Y_t)$ vanishes, and the objective collapses to the standard regression task $\mathbb{E}[Y_1|X]$. Conversely, in the zero noise limit $(t \to 0)$, the objective becomes an identity map that anchors the model directly to the data manifold. The intermediate regime teaches the model to  combine global guidance from $X$ with local geometry from $Y_t$.

The objective Eq.\eqref{eq:CDM_train_objective} statistical efficiency becomes clear when compared with alternative generative formulations like Conditional Flow Matching (CFM) \cite{Lipman2023}. The fundamental difference lies in the regression target. Our denoising model $g_\phi$ learns to predict $Y_1$. For a deterministic system, the conditional distribution $p_d(Y_1 |X)$ is delta-shaped, implying that the target has zero conditional variance.
In contrast, OT-based CFM model $v_\phi$ learns to predict the transport vector $Y_1 - Y_0$ from the interpolated state $Y_t = (1-t) Y_0 + t Y_1$ and condition $X$. The conditional distribution of this target, $(Y_1 - Y_0)|X$, remains a unit variance stochastic variable. This is because $Y_1|X$ is fixed, but $Y_0$ is an independent random variable, so $\text{Var}[(Y_1 - Y_0)|X] = \text{Var}[Y_1|X] + \text{Var}[Y_0] = 1$.

Crucially, the argument for statistical efficiency concerns the prediction {target}, not the input.
While exposing the model to noisy inputs $Y_t$ is essential for learning the global manifold geometry, including noise in the target (e.g., predicting velocity $Y_1 - Y_0$) forces the network to model high-entropy stochasticity~\cite{li2025basicsletdenoisinggenerative}.
By targeting the clean state $Y_1$ directly, our approach utilizes noise solely as a \textit{probe} to explore the state space.
This keeps the regression task deterministic, ensuring the model focuses entirely on resolving the physical signal rather than fitting random fluctuations.

\section{Experiments}
\label{sec:Experiments}

This section evaluates the proposed Conditional Denoising Models (CDM) on the Low-Temperature Plasma (LTP) benchmark.
Following a description of the experimental setup, architectures, and baselines, we present and analyze the results, assessing both predictive accuracy and physical adherence.

\subsection{Experimental Setup}

\textbf{Dataset and Preprocessing.} We evaluate our model on the low-temperature plasma dataset introduced by Valente et al. \cite{Valente2025}, generated using the LoKI\,\footnote{\url{https://nprime.tecnico.ulisboa.pt/loki/tools.html}} simulator \cite{Tejero-del-Caz2019, Tejero-del-Caz_2021}
The dataset comprises $3000$ samples mapping three scalar experimental controls (gas pressure, discharge current, and container radius) to a state vector of 17 chemical species densities. More details on the physical system can be found in \cite{Dias_2023}.

Prior to training, all input and output features are standardized (zero mean, unit variance) to ensure stable optimization dynamics. By default, the data is partitioned into a training set of $2550$ samples ($85\%$), a test set of $315$ samples ($10.5\%$), and a validation set of $135$ samples ($4.5\%$) for hyperparameter tuning.

\textbf{CDM-$t$ Architecture and Inference.} We adopt a multi-stream encoder-decoder architecture where the encoder independently processes three inputs: the physical condition $x$, the noisy state $\tilde{y}$, and the noise level $\sigma$. To handle the continuous nature of the time coordinate, $\sigma$ is projected using transformer-style sinusoidal positional embeddings \cite{Ho2020, Vaswani2023}. The resulting feature vectors are concatenated into a unified representation, which the decoder processes via layer normalization to regress the clean state $y$ concatenated with the input $x$. Layer configurations are detailed in Appendix \ref{appendix:PCDAE_architecture}.

For the inference phase, we analyze two different sampling strategies with a comparable computational budget of approximately $1300$ function evaluations. The first configuration, denoted as \textbf{CDM-$t$-dense}, employs a high-resolution noise schedule with $T=130$ levels and moderate refinement ($K=10$ steps per level), prioritizing a smooth trajectory along the diffusion path. In contrast, the second configuration, denoted as \textbf{CDM-$t$-sparse}, utilizes a coarse noise schedule with only $T=10$ levels, but performs more refinement ($K=130$ steps per level).

\textbf{CDM-$0$ Architecture and Inference.} To isolate the effect of time-dependency, the time-independent model shares the same architectural backbone as the time-dependent variant but removes the noise-embedding branch. This modification forces the encoder to learn a static, global vector field solely from the condition $x$ and the noisy state $\tilde{y}$. 

For inference, we evaluate the two fixed-point strategies defined in Algorithm \ref{alg:PCDAE_0_Inference}. We denote the model as \textbf{CDM-$0$-const} when using the fixed step size $\eta$ (implying geometric noise decay), and \textbf{CDM-$0$-adapt} when employing the adaptive step size based on the relative residual.

\textbf{Noise Schedule.} To ensure stable training and high-quality generation, we adopt a sine noise schedule adapted from \cite{nichol2021}. Unlike linear schedules, which can degrade performance by removing information too abruptly, the sine schedule maintains a smooth transition of noise levels. We parametrize the variance $\sigma^2(t)$ as:
\begin{equation}
    \sigma^2(t) = \sigma^2_{\max} \cdot \sin^2\left( \frac{t + s}{1 + s} \cdot \frac{\pi}{2} \right),
    \label{eq:noise_schedule}
\end{equation}
where $s=0.008$ \cite{nichol2021} and $\sigma_{\rm max}$ that we set as the default value of $\sigma_{\rm max} = 1$.

\textbf{Baseline Architectures.} We benchmark our physics-agnostic model against two physics-consistent baselines established in prior work \cite{Valente2025}. To ensure a rigorous comparison, both baselines replicate the exact neural architecture and training hyperparameters described in \cite{Valente2025}.

The first baseline is a \textbf{Physics-consistent Regressor (NN)}, a standard neural network trained to minimize a composite loss function, which enforces general physical laws (current conservation, the ideal gas law, and quasi-neutrality) alongside the data reconstruction error \cite{Valente2025}. The second baseline is the \textbf{Regressor with Projection (NN+Projection)}, which augments the physics-consistent regressor with a post-processing projection step \cite{Valente2025}. This additional stage strictly enforces the physical constraints during inference, converting the soft penalties of the loss function into hard geometric constraints.

\textbf{Evaluation Metrics.} We assess model performance using two complementary metrics: the standard Root Mean Square Error (RMSE) of the test set and the Physics RMSE. The former measures the Euclidean distance between the predicted chemical densities and the ground truth simulator outputs. The latter evaluates physical consistency by computing the RMSE of the constraint residuals $G(x, \hat{y})$, which represents the deviation from the physical laws (Charge Conservation, Ideal Gas Law, and Quasi-Neutrality). Ideally, this value should be zero, indicating strict adherence to the physical manifold. To ensure statistical robustness, we train an ensemble of models using 10 distinct train-test-validation splits.

\subsection{Results and Discussion}

We start by presenting the model and training set size efficiency comparison between the CDM variants (the time-independent CDM-$0$ ones and the time-dependent CDM-$t$ ones) and the baselines (NN and NN+Projection).
\begin{figure*}[h]
    \centering
    \includegraphics[width=1.0\linewidth]{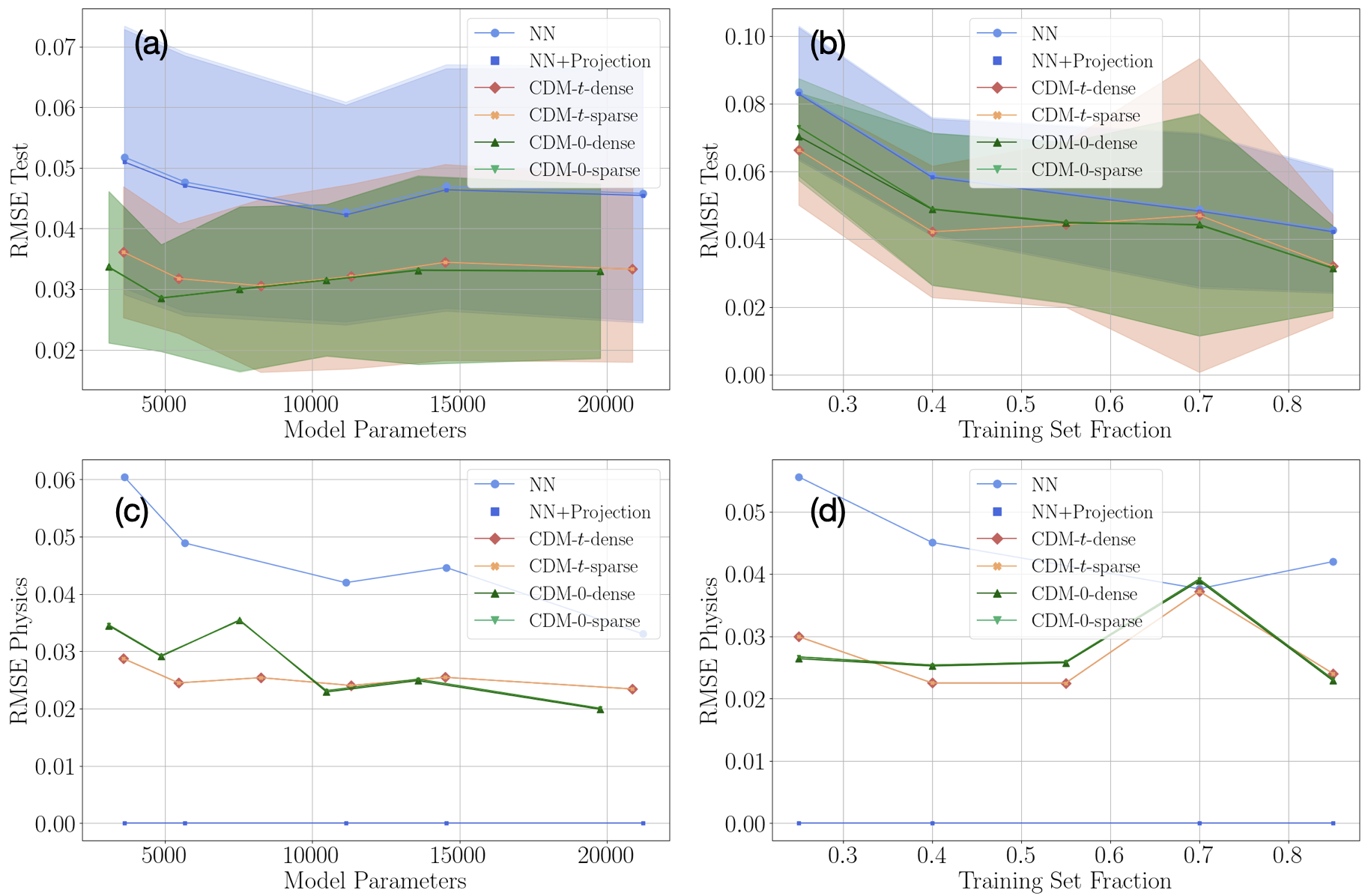}
    \caption{{Performance comparison between Physics-Consistent Baselines and Conditional Denoising Models (CDM).} 
    The top row {(a, b)} reports the predictive accuracy (Test RMSE), while the bottom row {(c, d)} evaluates physical consistency (Physics RMSE, measuring constraint violations). 
    The left column {(a, c)} analyzes the impact of model complexity (number of parameters), while the right column {(b, d)} evaluates robustness to data scarcity (training set fraction). 
    Shaded regions indicate the standard deviation across $10$ independent training runs. For the data-scarcity plots, the CDMs and baselines models have $\sim 15000$. Shaded regions represent the standard deviation across 10 distinct train-test-validation splits.}
    \label{fig:results_comparison}
    
\end{figure*}
Figure \ref{fig:results_comparison} presents a quantitative comparison between the physics-informed baselines (NN and NN+Projection) and the proposed Conditional Denoising Models (CDM). We analyze performance across two dimensions: model complexity (Panels \ref{fig:results_comparison}-(a), \ref{fig:results_comparison}-(c) and data scarcity (Panels \ref{fig:results_comparison}-(b), \ref{fig:results_comparison}-(d)).

Regarding predictive accuracy on the held-out test set (Panels \ref{fig:results_comparison}-(a) and \ref{fig:results_comparison}-(b)), we observe that all CDM variants consistently outperform the baselines across the entire spectrum of parameter counts. Notably, the time-independent CDM-$0$ variants achieve a significantly lower RMSE ($\approx 0.028$) with as few as $5000$ parameters, whereas the NN baselines stagnate around an RMSE of $0.045$ regardless of capacity. Furthermore, the CDM models exhibit substantially reduced variance (narrower shaded regions), suggesting that the denoising objective acts as a robust regularizer that stabilizes optimization compared to the supervised losses.

We also note that the specific inference strategies within each model class, specifically the comparison between constant and adaptive steps for CDM-$0$, and between dense and sparse schedules for CDM-$t$, yield very similar performance behaviors. This suggests that given a sufficient computational budget for inference (many steps), the generative flow robustly converges to the manifold regardless of the specific scheduling details.

Evaluating robustness to data scarcity (Panel \ref{fig:results_comparison}-(b)), we vary the training set fraction from $0.25$ ($\approx 800$ samples) to $0.85$ ($\approx 2550$ samples). The results indicate that the CDM variants generalize better in the low-data regime. Specifically, the CDM-$t$ variants demonstrate superior performance at the smallest sample sizes, suggesting that the continuous noise injection serves as an effective implicit data augmentation, allowing the model to capture the manifold geometry even when explicit training samples are sparse.

Finally, Panels~\ref{fig:results_comparison}-(c) and~\ref{fig:results_comparison}-(d) provide insight into physical consistency.
While the NN+Projection baseline achieves zero error by construction, the standard Physics-consistent NN exhibits the highest physics error ($\approx 0.04-0.06$) despite explicit training penalties.
In contrast, the CDM variants achieve consistently lower errors ($\approx 0.02-0.03$) without ever seeing the conservation equations.
This empirically validates the manifold hypothesis introduced before: rather than merely fitting a function, the CDM captures the topology of the solution subspace $\mathcal{M}$.
By learning to project points from the ambient space back onto the manifold, the model implicitly adheres to physical laws more effectively than standard regression constrained by soft penalties.

\begin{figure*}[h]
    \centering
    \includegraphics[width=1.0\linewidth]{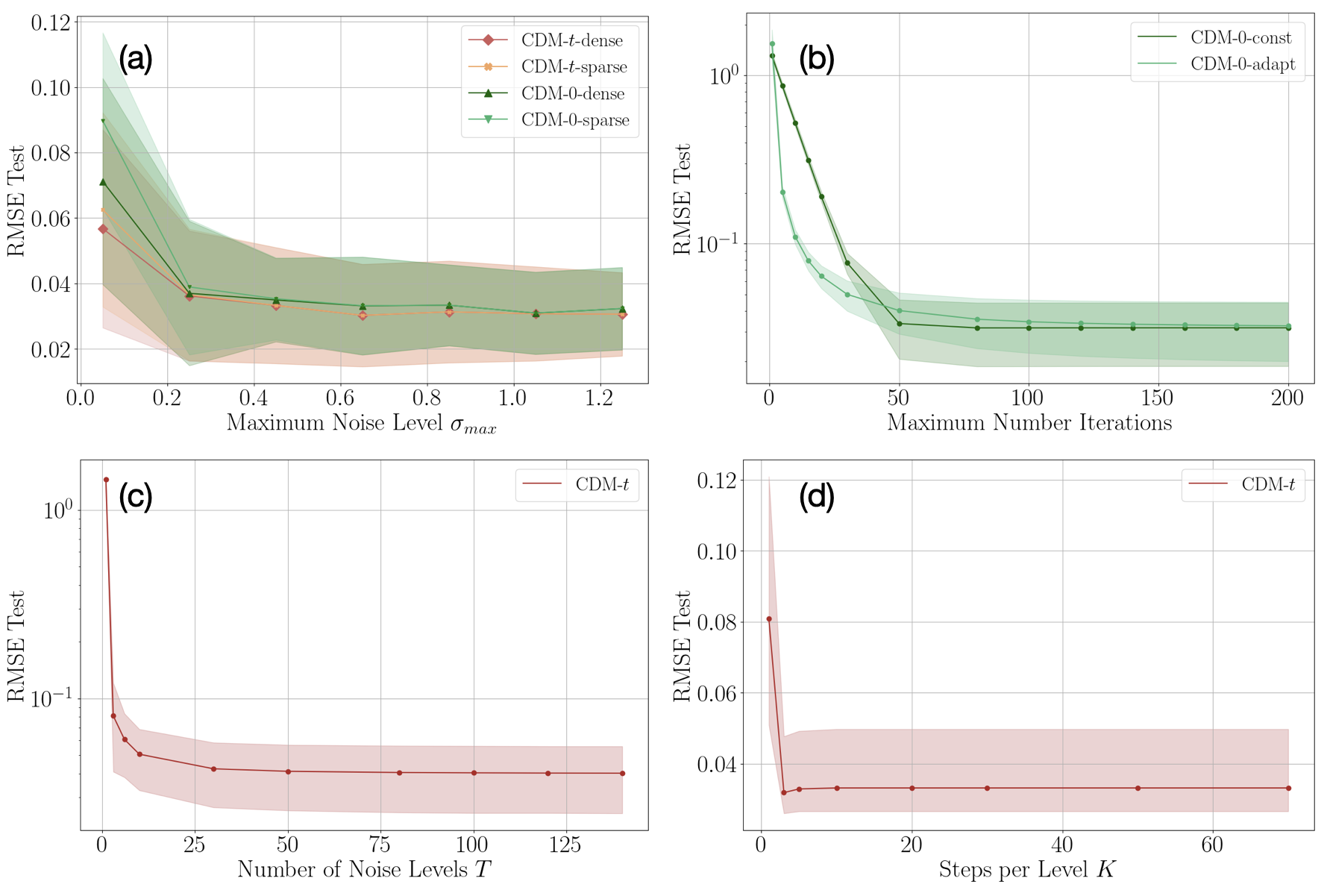}
    \caption{Ablation study on noise scaling and inference convergence. (a) Impact of the training noise scale $\sigma_{\max}$ on test accuracy (b) Convergence profile of time-independent CDM-$0$ models; the adaptive step size accelerates early-stage convergence, with both variants saturating around $50$--$80$ iterations. (c, d) Analysis of CDM-$t$ inference parameters. Panel (c) shows the effect of schedule resolution $T$ (with fixed $K=1$), saturating at $T \approx 30$. Panel (d) shows the impact of refinement steps $K$ (using a sparse schedule $T=3$). Shaded regions represent the standard deviation across 10 distinct train-test-validation splits. Note the logarithmic scale on the y-axis for Panels (b-c).}
    \label{fig:panel2}
    
\end{figure*}

Figure \ref{fig:panel2} presents an ablation study analyzing the sensitivity of the proposed models to the noise scale and inference hyperparameters.
First, we analyze the impact of the training noise scale $\sigma_{\max}$ in Panel \ref{fig:panel2}-(a). We observe that all model variants, both time-independent and time-dependent, exhibit nearly identical behaviors: performance is poor at low noise levels ($\sigma_{\max} < 0.2$) but improves rapidly and stabilizes once $\sigma_{\max} \geq 0.4$. This indicates that the corruption process must be sufficiently aggressive to cover the off-manifold neighborhood required for learning. Notably, the performance plateau beyond $\sigma_{\max}=0.4$ suggests that the model is robust to hyperparameter tuning; it does not degrade even when trained with excessive noise due to the strong guiding signal provided by the conditional input $x$.

Panel \ref{fig:panel2}-(b) illustrates the fixed-point convergence of the time-independent CDM-$0$ variants. Note the use of a logarithmic scale on the y-axis, highlighting the rapid error reduction. The curves reveal that the model effectively converges to the steady-state solution within approximately $50-80$ iterations. Moreover, the adaptive step-size variant exhibits a steeper initial descent compared to the constant-step variant, confirming that the adaptive mechanism effectively accelerates convergence in the early high-error phase.

For the time-dependent CDM-$t$ variants, we decouple the computational budget into the schedule resolution $T$ and the refinement depth per level $K$. Panel \ref{fig:panel2}-(c) shows the effect of increasing the schedule resolution $T$ while fixing the refinement to a single step per level ($K=1$). The results indicate that increasing the number of noise levels significantly improves accuracy up to a threshold of roughly $T \approx 50$, beyond which the benefits saturate. Conversely, Panel \ref{fig:panel2}-(d) analyzes the impact of local refinement $K$ using a very sparse schedule of only $T=3$ levels. The curve reveals that even with such a coarse schedule, the model can recover performance via intensive local refinement, saturating at approximately $K \approx 10$ steps per level.

A cross-comparison of these convergence profiles reveals a significant efficiency advantage. Unlike typical generative diffusion models for images that often require $\sim 1000$ sampling steps \cite{Ho2020}, our physical surrogates converge to the optimal solution manifold with a total budget of roughly $50-100$ function evaluations. We attribute this efficiency to two factors: the lower intrinsic dimensionality of the physical manifold compared to complex image spaces, and the strong conditioning signal provided by the inputs $x$, which effectively restricts the search space and guides the trajectory toward the solution.

\section{Conclusion}
\label{sec:Conclusions}

In this work, we introduce the {Conditional Denoising Model (CDM)}, a generative surrogate model designed to implicitly learn the geometry of physical solution manifolds in data-scarce regimes. By generalizing the classical Denoising Autoencoder to a continuous spectrum of noise levels, we study the gap between geometric manifold learning and diffusion-based generation. Crucially, we propose a time-independent formulation (CDM-$0$) that transforms the inference process from a reverse stochastic differential equation into a deterministic, iterative fixed-point refinement, acting as a learned contraction mapping towards the physical equilibrium.

Our experimental results on plasma physics simulations demonstrate that this approach offers superior parameter and data efficiency compared to physics-informed baselines. The CDM variants achieved a significant reduction in predictive error (RMSE $\approx 0.03$ vs. baseline $\approx 0.045$) even when trained on as few as $30\%$ of the available data ($1000$ samples). Furthermore, despite being physics-agnostic during training, the CDM implicitly captured the governing physical laws with greater fidelity than the physics-consistent neural network baseline, reducing constraint violations by approximately $50\%$. These findings validate that the denoising objective serves as a powerful implicit regularizer, allowing the model to internalize complex physical correlations without requiring explicit equation-based loss functions.

Looking forward, the results of the static CDM-0 suggest that time-independence is a robust and simpler inductive bias for steady-state problems. Future work will extend this finding by exploring energy-based formulations, where vector fields are derived from implicit potentials, and by scaling the framework to high-dimensional spatiotemporal domains.
Ultimately, this work opens new perspectives for the modeling of complex, steady-state physical systems, suggesting that implicitly learning the geometric manifold of solutions offers a robust and data-efficient alternative to explicit equation-based constraints.

\section*{Code Availability}
The source code implementing the Conditional Denoising Models (CDM) and the reproduction scripts are openly available at \href{https://github.com/joseAf28/CDM-PhysicsSurrogate}{https://github.com/joseAf28/CDM-PhysicsSurrogate}.

\section*{Acknowledgments}

This work was supported by the Portuguese FCT - Fundação para a Ciência e a Tecnologia, under project reference UID/50010/2023, UID/PRR/50010/2025 and LA/P/0061/2020, and by the European Union under Horizon Europe project CANMILK (DOI:https://doi.org/10.3030/101069491).
PV acknowledges support by project CEECIND/00025/2022 of FCT.

\bibliographystyle{unsrt}  
\bibliography{references}

\appendix

\section{Appendix}

\subsection{Derivation of the Conditional DCD Objective}
\label{app:DCD_derivation}

For completeness, we present the derivation of the objective in Eq.\eqref{eq:DCD_conditional_t}, adapting the formulation from \cite{luo2023} to the conditional setting. We assume the model's marginal $p_\theta(x)$ is fixed to the data marginal $p_d(x)$, allowing us to analyze the expected conditional KL divergence:
\begin{equation}
    \mathcal{D}_{\rm KL} (p_d(x, y) || p_\theta(x, y)) = \mathbb{E}_{x \sim p_d(x)} [\mathcal{D}_{\rm KL} (p_d(y|x) || p_\theta(y|x))].
\end{equation}

We start with the reconstruction objective (the expected conditional log-likelihood) and demonstrate its equivalence to the difference of KL divergences.
\begin{align*}
&\max_{\theta}\mathbb{E}_{p_d(x, y)} \mathbb{E}_{q_{t}(\tilde{y}|y,x)} \left[ \log p_\theta(y | \tilde{y}, x) \right] \\
&= \max_\theta \left( \mathbb{E}_{p_d(x, y)} \mathbb{E}_{q_{t}(\tilde{y}|y,x)} \left[ \log \frac{ p_\theta(y | \tilde{y}, x) p^{(t)}_\theta (\tilde{y}|x)}{ p_d(y | \tilde{y}, x) p^{(t)}_d (\tilde{y} |x)} \right] - \mathbb{E}_{p_d(x, y)} \mathbb{E}_{q_{t}(\tilde{y}|y,x)} \left[ \log \frac{ p^{(t)}_\theta (\tilde{y}|x)}{ p^{(t)}_d (\tilde{y} | x)} \right] \right) \\
& \text{\small (Using Bayes' rule: $p(y|\tilde{y},x)p^{(t)}(\tilde{y}|x) = q_t(\tilde{y}|y,x)p(y|x)$ for both $p_d$ and $p_\theta$)} \\
&= \min_\theta \left( \mathbb{E}_{p_d(x,y)} \mathbb{E}_{q_t(\tilde{y}|y,x)} \left[ \log \frac{ p_d(y|x) q_{t} (\tilde{y}|y,x)}{ p_\theta(y|x) q_{t} (\tilde{y}|y,x)} \right] - \mathbb{E}_{p_d(x, y)} \mathbb{E}_{q_{t}(\tilde{y}|y,x)} \left[ \log \frac{ p^{(t)}_\theta (\tilde{y}|x)}{ p^{(t)}_d (\tilde{y} | x)} \right] \right) \\
& \text{\small (Canceling $q_t$ and noting the first term is independent of $\tilde{y}$)} \\
&= \min_\theta \left( \mathbb{E}_{p_d(x,y)} \left[ \log \frac{ p_d(y|x)}{ p_\theta(y|x) } \right] - \mathbb{E}_{p_d(x, y)} \mathbb{E}_{q_{t}(\tilde{y}|y,x)} \left[ \log \frac{ p^{(t)}_\theta (\tilde{y}|x)}{ p^{(t)}_d (\tilde{y} | x)} \right] \right) \\
& \text{\small (Identifying the terms as expectations of KL divergences)} \\
&= \min_\theta \mathbb{E}_{p_d(x)} \left[ \mathcal{D}_{\rm KL}(p_d(y|x) || p_\theta(y|x)) - \mathcal{D}_{\rm KL}(p^{(t)}_d(\tilde{y}|x) || p^{(t)}_\theta(\tilde{y}|x)) \right] \\
&= \min_\theta \mathbb{E}_{p_d(x)} \left[ \mathcal{D}^{(t)}_{\rm DCD} (p_d(y|x) || p_\theta(y|x)) \right]
\end{align*}
where we used the definition $p^{(t)}(\tilde{y}|x) = \int dy \, q_t(\tilde{y} | y) p(y|x)$.
Thus, maximizing the reconstruction objective is equivalent to minimizing the conditional DCD divergence.

\subsection{Relation between the Optimal Denoiser and Conditional Score}
\label{app:vector_field_and_conditional_score}

In this section, also for completeness, we present the derivation the conditional version of \textit{Tweedie}'s formula \cite{Park2025_tweedie}. We show that the optimal denoiser $g_{\phi^*}$ is related to the score of the conditional marginal perturbed data distribution  $\nabla_{\tilde{y}} \log q_t(\tilde{y} | x)$.

We assume the forward process is defined as $q_t(\tilde{y} | y, x) = \mathcal{N}(\tilde{y}; y, \sigma^2(t)I)$. This implies $\tilde{y} = y + \sigma(t)\epsilon$ for $\epsilon \sim \mathcal{N}(0, I)$. As a result, we can find the gradient of the log-likelihood:
\begin{equation}
    \nabla_{\tilde{y}} \log q_t(\tilde{y} | y, x) = \nabla_{\tilde{y}} \left( - \frac{1}{2\sigma^2(t)} \|\tilde{y} - y\|^2_2 \right) = - \frac{1}{\sigma^2(t)}(\tilde{y} - y) = \frac{y - \tilde{y}}{\sigma^2(t)}.
\end{equation}

We begin with the definition of the conditional score $s^*(\tilde{y} , x)$ and show its relation to $g_{\phi^*}(\tilde{y}, x, t)$:
\begin{align*}
s^*(\tilde{y} , x) &= \nabla_{\tilde{y}} \log q_t(\tilde{y} | x) = \frac{1}{q_t(\tilde{y} | x)} \nabla_{\tilde{y}} q_t(\tilde{y} | x)  = \frac{1}{q_t(\tilde{y} | x)} \nabla_{\tilde{y}} \int dy \, p(y|x) q_t(\tilde{y} | y, x) \\
&= \frac{1}{q_t(\tilde{y} | x)} \int dy \, p(y|x) \left( q_t(\tilde{y} | y, x) \nabla_{\tilde{y}} \log q_t(\tilde{y} | y, x) \right) \\
&= \frac{1}{q_t(\tilde{y} | x)} \int dy \, p(y|x) q_t(\tilde{y} | y, x) \left( \frac{y - \tilde{y}}{\sigma^2(t)} \right) \quad \text{\small (Using Gaussian assumption)} \\
&= \int dy \, \frac{p(y|x) q_t(\tilde{y} | y, x)}{q_t(\tilde{y} |x)} \left( \frac{y - \tilde{y}}{\sigma^2(t)} \right) = \int dy \, q(y | \tilde{y}, x, t) \frac{y - \tilde{y}}{\sigma^2(t)} \quad \text{\small (By conditional Bayes' rule)} \\
&= \frac{1}{\sigma^2(t)} \mathbb{E}_{q(y | \tilde{y}, x, t)} [y - \tilde{y}] = \frac{1}{\sigma^2(t)} \mathbb{E} [y - \tilde{y} | \tilde{y}, x, t].
\end{align*}
Since the optimal denoiser corresponds to $g_{\phi^*}(\tilde{y}, x, t) = \mathbb{E}[y | \tilde{y}, x, t]$, we arrive at the following formula:
\begin{equation}
    g_{\phi^*}(\tilde{y}, x, t) = \mathbb{E}[y | \tilde{y},x, t] = \tilde{y} +\sigma^2(t) \nabla_{\tilde{y}} \log q_t(\tilde{y}|x).
\end{equation}


\subsection{CDM Model Architecture and Default Hyperparameters}
\label{appendix:PCDAE_architecture}

The detailed layer-by-layer specification of our CDM-$t$ model is provided in Table \ref{tab:architecture}.

\begin{table}[h!]
    \centering
    \caption{Layer-wise architecture and default dimensions of the CDM-$t$ model. $B$ denotes the batch size.}
    \label{tab:architecture}
    \begin{tabular}{llc}
        \toprule
        \textbf{Module} & \textbf{Operation} & \textbf{Output Dimension} \\
        \midrule
        \multicolumn{3}{c}{\textit{--- Inputs ---}} \\
        Input Y & Noisy State $\tilde{y}$ & $(B, \#y)$ \\
        Input X & Input Condition $x$ & $(B, \#x)$ \\
        Input $\sigma$ & Noise Level $\sigma$ & $(B, 1)$ \\
        \midrule
        \multicolumn{3}{c}{\textit{--- Encoder / Projections ---}} \\
        Noise Embedding & Sinusoidal Emb. + 2-Layer MLP & $(B, 10)$ \\
        Encoder Y Path & 2-Layer MLP (GELU) & $(B, 58)$ \\
        Encoder X Path & 2-Layer MLP (GELU) & $(B, 16)$ \\
        Fusion & Concatenate [Enc(Y), Enc(X), Emb($\sigma$)] & $(B, 84)$ \\
        \midrule
        \multicolumn{3}{c}{\textit{--- Decoder ---}} \\
        Pre-processing & Layer Normalization & $(B, 84)$ \\
        Main Block & Linear $\rightarrow$ GELU & $(B, 58)$ \\
        Output Head & Linear Layer (predicts $y$ and $x$) & $(B, \#y + \#x)$ \\
        \bottomrule
    \end{tabular}
\end{table}

To ensure stable convergence, we enforce a minimum training duration of $4500$ epochs before activating the stopping criteria. Beyond this point, we employ Early Stopping monitoring the validation loss with a patience of $20$ epochs.
The default hyperparameters used for model training and inference are detailed in Table \ref{tab:training_params}.

\begin{table}[h!]
    \centering
    \caption{Default training hyperparameters.}
    \label{tab:training_params}
    \begin{tabular}{lc}
        \toprule
        \textbf{Parameter} & \textbf{Value} \\
        \midrule
        \multicolumn{2}{c}{\textit{Noise Configuration}} \\
        \midrule
        Noise Kernel ($q_t(\cdot \mid y)$) & Isotropic Gaussian $\mathcal{N}(y, \sigma(t)^2 \mathbf{I})$ \\
        Noise Schedule ($\sigma(t)$) &  sine schedule Eq.\eqref{eq:noise_schedule}\\
        Max Noise Level ($\sigma_{\text{max}}$) & 1.0 \\
        CDM-$0$-const step size ($\eta$) & 0.1 \\
        CDM-$0$-adapt base step size ($\eta_{\rm base}$) & 1.0  \\
        CDM-$0$ ($N_{\rm max}$) & 1300 \\
         CDM-$0$ ($\epsilon_{\rm conv}$) & $10^{-5}$ \\

        \midrule
        \multicolumn{2}{c}{\textit{Optimizer Configuration}} \\
        \midrule
        Optimizer & Adam \cite{Kingma2017} \\
        Learning Rate & $1 \times 10^{-3}$ \\
        Batch Size & 128 \\
        Min Number Training Epochs & 4500 \\
        \bottomrule
    \end{tabular}
\end{table}


\end{document}